\def\year{2022}\relax
\documentclass[letterpaper]{article} 
\usepackage{aaai22}  
\usepackage{times}  
\usepackage{helvet}  
\usepackage{courier}  
\usepackage[hyphens]{url}  
\usepackage{graphicx} 
\usepackage{newfloat}
\usepackage{listings}
\usepackage{amssymb}
\usepackage{fontawesome5} 
\usepackage[most]{tcolorbox}
\usepackage{lipsum} 
\usepackage{natbib}  
\usepackage{caption} 
\DeclareCaptionStyle{ruled}{labelfont=normalfont,labelsep=colon,strut=off} 
\usepackage{algorithm}
\usepackage{algorithmic}
\usepackage{amsmath}
\usepackage{tabularray}
\usepackage{newfloat}
\usepackage{listings}
\floatstyle{ruled}
\newfloat{listing}{tb}{lst}{}
\floatname{listing}{Listing}

\title{MALLES: A Multi-agent LLMs-based Economic Sandbox with Consumer Preference Alignment}
\author {
    Yusen Wu,
    Yiran Liu,
    Xiaotie Deng
}
\affiliations {
}

\usepackage{bibentry}

\begin{document}

\maketitle

\begin{abstract}
  In the real economy, modern decision-making is fundamentally challenged by high-dimensional, multimodal environments, which are further complicated by agent heterogeneity and combinatorial data sparsity. This paper introduces a Multi-Agent Large Language Model-based Economic Sandbox (MALLES), leveraging the inherent generalization capabilities of large-sacle models to establish a unified simulation framework applicable to cross-domain and cross-category scenarios. Central to our approach is a preference learning paradigm in which LLMs are economically aligned via post-training on extensive, heterogeneous transaction records across diverse product categories. This methodology enables the models to internalize and transfer latent consumer preference patterns, thereby mitigating the data sparsity issues prevalent in individual categories. To enhance simulation stability, we implement a mean-field mechanism designed to model the dynamic interactions between the product environment and customer populations, effectively stabilizing sampling processes within high-dimensional decision spaces. Furthermore, we propose a multi-agent discussion framework wherein specialized agents collaboratively process extensive product information. This architecture distributes cognitive load to alleviate single-agent attention bottlenecks and captures critical decision factors through structured dialogue. Experiments demonstrate that our framework achieves significant improvements in product selection accuracy, purchase quantity prediction, and simulation stability compared to existing economic and financial LLM simulation baselines. Our results substantiate the potential of large language models as a foundational pillar for high-fidelity, scalable decision simulation and latter analysis in the real economy based on foundational database.
\end{abstract}

\section{Introduction}

The real-economy sectors and retail industries have undergone rapid digitization, catalyzing a paradigm shift toward intelligent enterprise operations. This transformation has necessitated the adoption of dynamic algorithms and rule-based systems to optimize decision-making processes. Prominent methodologies include deep reinforcement learning \cite{deep-rl-strategy}, Growth Accounting frameworks \cite{backward}, generative supply chains \cite{collaboration-paradox}, symbolic regression \cite{deeprule}, and agentic AI for Autonomous Business Models \cite{transform-invest}. These approaches are predicated on data-driven expansion and the precise prediction of market behaviors, with a specific emphasis on consumer preference modeling.

Historically economic simulation systems have modeled these behaviors through techniques such as rule summarization \cite{portrait}, hybrid choice models \cite{novel-products}, and macro-level forecasting integrated with global events \cite{world-event}. Other established methods encompass deep learning-based demand analysis \cite{demand-analysis}, Neuro-Fuzzy Modelling \cite{neuro-fuzzy}, linear demand implementations \cite{linear-demand}, survey-based forecasting\cite{surveys}, and time series analysis \cite{time-series}. While foundational, these approaches are frequently constrained by task-specificity and the necessity for manual feature engineering. Recently, Generative Large Language Models (LLMs) have been deployed to simulate market rules and consumer behaviors by leveraging textual data \cite{llm-cpi, llm-prediction-lab, decoding}. Nevertheless, existing simulation architectures encounter critical bottlenecks in preference prediction: (1) Data sparsity at the category level, where sparse transaction records fail to span the combinatorial feature space; (2) Poor Out-Of-Distribution (OOD) generalization, limiting applicability to novel or long-tail categories; and (3) High dimensionality of product characteristics, where unstructured textual and multimodal attributes challenge traditional modeling efficacy\cite{abides-economist, eco-agent}.

To address these systemic impediments, we introduce MALLES, a framework that harnesses the generalization and alignment capabilities of LLMs to simulate consumer preferences derived from historical transaction data. MALLES utilizes cross-category transaction records via post-training, enabling the system to abstract and transfer underlying preference patterns, thereby ameliorating data sparsity and enhancing OOD performance\cite{llm-as-eco-agents-home-silicus}. Unlike conventional deep learning paradigms that necessitate fixed-length inputs and manual normalization\cite{eco-agent, abides-economist}, MALLES leverages the flexible input context and inherent textual generalization of LLMs, facilitating the efficient processing of heterogeneous product information\cite{llm-as-eco-agents-home-silicus, multi-agent-public-analysis}.

The core technical architecture of MALLES incorporates three strategic mechanisms. First, it employs a multi-agent discussion mechanism to distribute cognitive load when processing long-context product information; this compresses high-dimensional data into salient decision factors through structured dialogue, alleviating attention bottlenecks and augmenting interpretability \cite{improve-negotiate, agent-exchange}. Second, MALLES integrates mean-field stabilization to model the dynamic interplay between product environments and customer populations, ensuring robust sampling within high-dimensional decision spaces \cite{multi-agent-public-analysis}. Third, the framework utilizes input augmentation with attention control to reconstruct partial observations, thereby reducing biases arising from hidden variables. Extensive experiments demonstrate that MALLES achieves statistically significant improvements in product selection accuracy, purchase quantity prediction, and simulation stability compared to existing economic and financial LLM baselines \cite{eco-agent, abides-economist, fincon}. These results underscore the potential of MALLES as a foundation for high-fidelity, scalable decision simulation in real-economy scenarios.

The primary contributions of this work are summarized as follows: (1) The development of MALLES, a unified LLM-based economic sandbox that leverages heterogeneous data to generalize consumer preferences across categories, effectively overcoming data sparsity and OOD generalization challenges; (2) A proprietary multi-agent discussion framework designed to process high-dimensional product contexts, which enhances simulation stability and interpretability through collaborative reasoning; and (3) The implementation of comprehensive stabilization mechanisms—specifically mean-field modeling and attention control—to ensure robust performance in complex, high-dimensional decision environments.

\section{Related Work}

\textbf{LLM-Driven Economic Agent Simulation. }
Accurate consumer preference prediction is crucial for optimizing product selection, inventory management, and marketing strategies in digitally transformed real economy. Traditional simulation systems, however, face challenges such as category-level data sparsity, poor out-of-distribution (OOD) generalization, and inefficiencies in handling high-dimensional product features. As product representations evolve from categorical to multimodal and textual descriptors, conventional methods relying on manual feature engineering and deep learning suffer from information loss and computational inefficiency.

Recent research has explored LLMs as economic agents for simulating market behavior. Integrating agent-based modeling (ABM) with LLMs has emerged as a promising direction. Frameworks like \textit{EconAgent}, \textit{LLM Economist}, and \textit{ABIDES-Economist} construct heterogeneous agents through persona-conditioned prompts and memory modules to simulate micro- and macro-level market phenomena \cite{eco-agent, llm-as-eco-agents-home-silicus, abides-economist}. These works show LLMs’ ability for rich semantic interactions in policy evaluation and market design \cite{finmem}. However, they primarily focus on macro-level reasoning and semantic interactions, neglecting \textit{numerical sensitivity} (e.g., price elasticity) and effective multimodal alignment \cite{empowering-eco-sim}. Moreover, these approaches typically rely on pretrained models without optimization for real transaction data, limiting generalization in data-sparse scenarios.

\textbf{Financial and Trading LLM Architectures. }
In finance, works like \textit{FinMem}, \textit{FinCon}, and \textit{InvestorBench} apply LLMs to trading and risk management, introducing hierarchical memory structures, workflow management, and backtesting mechanisms \cite{finmem, fincon, INVESTORBENCH}. These frameworks provide engineering insights for risk control and profit evaluation, relevant to wholesale decisions involving profit maximization and inventory turnover \cite{greedllama}. While laying foundations for automated procurement prediction, these methods target at financial market tasks and lack integration with natural language dialogue and negotiation in retail/wholesale contexts. Consequently, they struggle to support interpretable decision formula discovery and exhibit limited capability in handling multimodal inputs and cross-category generalization, hindering applicability in real economy scenarios.

\textbf{Economic Simulation Benchmarks.} To assess economic rationality, benchmarks like \textit{GLEE}, \textit{EconArena}, and \textit{STEER} propose language-based economic games and rationality metrics including consistency, sensitivity, and efficiency \cite{glee, steer, stockbench}. Complementary research enhances negotiation quality via self-play and AI feedback \cite{improve-negotiate}. These studies lay groundwork for dialogue-driven marketing and agent evaluation. However, existing benchmarks emphasize semantic-level rationality while inadequately addressing numerical sensitivity and multimodal alignment evaluation. They also lack comprehensive testing of OOD generalization in data-sparse environments.

\textbf{Research Gaps and Our Contributions.} In summary, existing high-fidelity economic simulation research faces significant limitations: (1) most LLM-agent frameworks inadequately address numerical sensitivity and multi-dimensional alignment, hindering accurate simulation of behaviors like price responsiveness; (2) current approaches struggle to handle data sparsity and generalize to new or OOD categories. To address these issues, we propose a unified LLM-based economic sandbox, featuring two key innovations: cross-category data post-training for numerical alignment, and multi-agent discussion and mean-field mechanisms.  These innovations enable significant improvements in prediction accuracy and simulation stability in data-constrained real economy scenarios.

\section{Preliminaries: LLM-Empowered Economic Agent Simulation}

\textbf{EconAgent}~\cite{eco-agent} introduces a macroeconomic simulation framework that leverages LLMs to generate agents with dynamic decision-making. It constructs heterogeneous agents through perception (realistic profiles), memory (multi-period experiences), and action (LLM-generated work/consumption propensities based on economic prompts) modules. While successfully replicating macroeconomic phenomena, it focuses on high-level household decisions with relatively simple prompting. Our work extends this paradigm to \textit{microeconomic} retail/wholesale scenarios incorporating alignment techniques via transaction data post-training and multi-agent dialogue frameworks.

The \textbf{ABIDES-Economist} simulator~\cite{abides-economist} models heterogeneous agents using multi-agent reinforcement learning (MARL) within a Partially Observable Markov Game. Household agents optimize consumption and labor decisions, while firm agents learn price and wage policies. For tractability, it employs shared policy networks per agent type and tiered learning rates. This approach generates emergent behaviors validated against macroeconomic stylized facts, providing a robust but computationally intensive foundation for agent-based economic simulation.

The \textbf{FinCon} framework~\cite{fincon} simulates economic agent behavior in financial institutions through a multi-agent LLM architecture. It deploys specialized analyst agents for various data modalities (text, audio, quantitative data) and a manager agent that synthesizes insights and updates investment beliefs. Economic behavior is refined via within-episode risk control and a conceptual verbal reinforcement mechanism for cross-episode learning. While this setup mimics organizational learning in investment firms, it is primarily designed for financial market tasks and lacks direct integration with retail/wholesale negotiation processes.

The \textbf{LLM Economist} framework~\cite{llm-as-eco-agents-home-silicus} simulates economic agents within a hierarchical Stackelberg game. Worker agents, conditioned with persona-specific prompts, optimize labor supply through in-context natural language reasoning. A planner agent uses in-context reinforcement learning to propose tax schedules aimed at maximizing social welfare. The framework incorporates democratic voting to influence policy evolution, modeling institutional dynamics through language-driven interactions.

\section{Methodology}
\label{sec:methodology}

\subsection{Framework Overview}

Modern enterprise decision-making in real economy is heavily reliant on accurate assessing consumer responses to pricing, promotion, and product strategies. Traditional simulation systems struggle with consumer preference modeling due to data sparsity at the category-level data sparsity, poor out-of-distribution generalization, and inefficient handling of high-dimensional multimodal product features. To address these limitations, we introduce a Multi-Agent Large Language Model-based Economic Sandbox (MALLES). It utilizes cross-category transaction data and collaborative multi-agent reasoning to simulate heterogeneous consumer behaviors with high fidelity.

The framework adopts a hierarchical decision-making paradigm, where strategic objectives and constraints propagate top-down through organizational layers. Upper management sets broad targets, which are decomposed into feasible sub-targets for lower-level optimization. These targets are validated within the economic sandbox, and successful implementations are adopted, while identified bottlenecks trigger recursive feedback for adjustment. This closed-loop process iteratively refines both strategic objectives and underlying decision parameters, with LLMs dynamically updating parameters based on experimental feedback and emerging patterns.

A key innovation is the use of large, heterogeneous transaction records to economically align LLMs via specialized post-training. This enables effective knowledge transfer across categories and mitigates the issue of per-category data sparsity. We distinguish between retail and wholesale simulations: retail customers focus on need satisfaction and price comparisons aligned with consumption values, whereas wholesale customers engage in profit-driven decisions that consider market dynamics, inventory turnover, and clearance mechanisms. By unifying both paradigms, MALLES provides a testing environment for real-world decision optimization, bridging the gap between theoretical models and practical business applications.

\subsection{Problem Formulation and Notation}

Consider a simulation environment where real decision makers (retail customers, wholesalers, manufacturers, etc.) are indexed by \(i\). Simulated decisions from agent models are denoted \(\hat{a}_i\), and actual observed behaviors are \(a_i\). Observable input information is \(X_i^{obs}\), while hidden factors are \(X_i^{hid}\). Personality, preference, and behavioral style parameters are \(\rho_i\). We assume a true decision function governs real behavior:$a_i = \mathcal{D}(X_i^{obs}, X_i^{hid}, \rho_i)$

The simulation agent decides via $\hat{a}_i = \hat{\mathcal{D}}(X_i^{obs}, Z_i, \hat{\rho}_i; \theta)$ where \(Z_i\) is a constructed profile summary approximating \(X_i^{hid}\), \(\hat{\rho}_i\) are agent personality/preference parameters for style calibration, and \(\theta\) encompasses all trainable parameters, including LLM components and specialized adaptive modules. The overall error objective is the expected loss between simulated and actual decisions $\mathcal{E} = \mathbb{E}_i[\ell(a_i, \hat{a}_i)]$, where \(\ell\) is a suitable loss function (e.g., squared error for continuous decisions, classification error for discrete choices, or profit differential for business outcomes). Our modular design ensures \(\mathcal{E}\) decreases progressively through training and simulation while maintaining interpretable decision structures and strong generalization across diverse economic scenarios.

\subsection{Retail and Wholesale Customer Simulation with Economic Alignment and Multi-Agent Discussion}

Retail customer simulation centers on training LLMs to be economically aware via post-training on extensive product-purchase records, cultivating a monetary consciousness aligned with real consumer behavior. Unlike traditional methods constrained by domain-specific data scarcity, our approach leverages cross-category transaction data to mitigate per-category sparsity. Given customer demand \(d\), candidate product information \(p\) with multimodal features, and discount incentives \(c\), the LLM iteratively generates decision functions to yield purchase intent and quantity. Specialized attention mechanisms in the prompt architecture ensure appropriate sensitivity to numerical economic factors, so prices, discounts, and other quantitative elements are properly weighed.

To handle partial observability where hidden variables affect outcomes, we employ input augmentation. A profile summarization module constructs long-term customer profiles \(Z_i\) capturing historical purchase patterns, category preferences, promotion sensitivity, brand affinities, and communication styles. These summaries allow the simulation input \((X_i^{obs}, Z_i)\) to approximate the full decision context. Additionally, attention priors and weighting controls prioritize price and promotion features within multimodal inputs, enforced via regularization and an attention matching loss: $\mathcal{L}_{attn} = \mathbb{E}_i[\mathrm{KL}(A_i \| A_i^*)]$ where \(A_i\) is the actual attention distribution and \(A_i^*\) the prior emphasizing economic features, ensuring stable and rational attention during simulation. The retail model \(\pi_{\text{retail}}\) processes input \(x_{\text{retail}}\) through iterative reasoning. Its raw output is parsed by \(\phi_{\text{retail}}\) to extract structured decisions $a_{\text{retail}} = (b_r, q_r) = \phi_{\text{retail}}(\pi_{\text{retail}}(x_{\text{retail}}))$ with \(b_r \in \{0,1\}\) for purchase decision and \(q_r\) for quantity. When customer data is limited, historical purchase records serve as baselines, shifting the training objective toward predicting individual styles from context rather than averaging population behavior.

Correspondingly wholesale decisions involve substantial financial commitments and depend on profitable inventory turnover, requiring analysis of market dynamics, price evolution, and clearance mechanisms. We thus employ a multi-agent framework to derive profit-oriented purchasing formulas through LLM adjustment combined with numerical optimization. The architecture includes three roles: wholesalers, marketing personnel (customer service), and manufacturers. Using conversation logs, we do agent characterizations with collective query formulation and dialogue labeling. The framework enables multi-round dialogue games to refine purchasing formulas obtained via symbolic regression.

Simulation proceeds through structured multi-round dialogue. Starting with background \(b = x_{\text{wholesale}}\) and history \(h_0 = \{b\}\), in round 1 the dealer agent \(\pi_{\text{dealer}}\) produces analysis \(a_1\) from \(h_0\) (competitive context, historical performance, objectives), updating to \(h_1 = h_0 \cup \{(\text{dealer}, a_1)\}\). In round 2, the service agent \(\pi_{\text{service}}\) responds with analysis \(a_2\) addressing dealer points and providing promotions, yielding \(h_2 = h_1 \cup \{(\text{service}, a_2)\}\). Round 3 has the manufacturer agent \(\pi_{\text{manufacturer}}\) contribute analysis \(a_3\) on production constraints, supply, and partnerships, producing \(h_3 = h_2 \cup \{(\text{manufacturer}, a_3)\}\). Rounds \(4\) to \(n-1\) involve iterative discussion, building on prior analyses and updating the history. In the final round \(n\), the dealer agent synthesizes the discussion into analysis \(a_n\), updating to \(h_n = h_{n-1} \cup \{(\text{dealer}, a_n)\}\). Finally the parsing function \(\phi_{\text{wholesale}}\) extracts the purchasing decision $(s_w, q_w) = \phi_{\text{wholesale}}(h_n)$ where \(s_w\) denotes product selection and \(q_w\) for purchase quantity. To enhance interpretability and economic rationality symbolic regression is integrated within the multi-agent framework to discover compact mathematical expressions of decision patterns. Agents collectively reason about economic factors and their relationships, generating human-understandable rules that align with theory while fitting observed transaction data.

\subsection{Stabilization and Calibration Mechanisms}

Despite input augmentation, differences in thinking styles between artificial agents and humans may persist. We address this by introducing style parameters \(\rho_i\) and specialized submodules that integrate them into cognitive architectures. A behavioral economics component injects commercial insights (e.g., discount sensitivity, loss aversion) as elements of \(\rho_i\) into prompts or adapters to align decisions with human preference patterns. A hybrid cognitive architecture fuses cognitive systems (e.g., ACT-R) with LLMs. A cognitive core module \(C\) generates candidate strategy sets from \(Z_i\) and historical decisions while the LLM semantically understands and scores these, producing \(\hat{a}_i\). This stabilizes rule-based or heuristic decision structures, with \(\rho_i\) weighting candidate generation versus scoring.

Consistency regularization and multi-sampling further stabilize decisions. Multiple responses \(\hat{a}_{i,1}, \hat{a}_{i,2}, \dots\) are generated for the same \((X_i^{obs}, Z_i, \rho_i)\) under minor perturbations (e.g., small price changes). A consistency loss reduces output randomness $\mathcal{L}_{cons} = \mathbb{E}_i \mathbb{E}_{\epsilon \sim \mathcal{N}(0,\sigma)}[|\hat{a}_i(X_i^{obs} + \epsilon, Z_i, \rho_i) - \hat{a}_i(X_i^{obs}, Z_i, \rho_i)|^2]$. Since individual errors may accumulate into population-level deviations, we implement statistical population-level correction to reduce overall error \(\mathcal{E}\) and enhance generalization. A mean-field alternating update mechanism creates interactions between micro-level strategy generation and macro-level population distributions. Iteratively, the model’s output distribution of customer responses approximates the real distribution. Each iteration \(t\) maintains a macro-response variable \(\mu_t\) that influences agent inputs (e.g., market context), and agent outputs combine to form \(\mu_{t+1}\).

Given real datasets \(\{a_i\}_{i\in I_{\text{real}}}\) with inputs \(\{X_i^{obs}, Z_i\}\), we estimate conditional distribution \(P_{\text{real}}(a|X,Z)\) and simulation distribution \(P_{\text{sim}}(a|X,Z)\). Calibration uses mapping functions \(f\) and reweighting factors \(w(X,Z)\) via $\min_f D_{\text{KL}}(P_{\text{real}}(a|X,Z) \| f(P_{\text{sim}}(a|X,Z)))$. Alternatively, Wasserstein distance may be used. The calibration function \(f\) can post-process outputs or adjust prompts/preference parameters. Conflict feedback and bottleneck detection enhance robustness. If key indicators (margins, inventory costs, sales distributions) deviate beyond threshold \(\delta\) from historical patterns or macro targets, feedback signals \(B\) trigger re-adjustment of higher-level targets, style parameters, or price constraints. This process incorporates redundancy: initial target flexibility allows bottleneck identification at lower levels, with targets tightening over iterations to achieve convergence.

\subsection{Framework Implementation}

MALLES integrates multimodal, multi-source data via standardized schemas. In pratice, product information includes IDs, categories, base prices, image embeddings/files, attributes, and historical sales series. Transaction records form the core dataset, with rows for individual orders containing timestamps, customer IDs and types, product IDs, quantities, unit prices, discounts, channels, and review scores. When available, dialogue logs capture interactions via dialogue IDs, participant roles, timestamped turns, and negotiated outcomes. Customer profiles include income brackets, buyer types, and historical purchase profiles. Statistical inventory turnover and decision inertia enhance realism by computing stable inventory baselines from regularity patterns in retail/wholesale contexts. Price distributions affect final category decisions with measurable inertia, captured via weighting schemes—manifesting as brand switching in retail and proportional brand allocation in wholesale.

The framework combines economic multi-agent role-playing with numerical alignment in data-rich but sample-sparse settings, serving both decision analysis and intelligent marketing training. Multi-agent approaches improve alignment through greater reasoning stochasticity versus single‑agent thinking, while multi‑agent reinforcement learning increases data efficiency via reward shaping. Properly aligned multi‑agent systems better approximate real decision contexts, offering advantages over costly traditional evaluations, error‑prone deep‑learning sandboxes with limited data, and existing LLM‑agent simulations with poor data sensitivity and distribution misalignment. Unlike traditional role‑playing which relies on extensive fine‑tuning and manual prompt design for tonal styles, our decision‑oriented methodology characterizes concrete behaviors, leveraging the logical reasoning of multi‑agent LLMs. Integrating role‑playing with symbolic regression harnesses semantic understanding to mitigate product data sparsity while incorporating numerical sensitivity, establishing a robust basis for high‑fidelity simulation in high‑dimensional, multimodal, heterogeneous real‑economy environments.

\subsection{Theoretical Guarantees}

We present the main theorem that quantifies the advantage of our cross-category training paradigm. For a target category $c_t$ with limited data, the generalization improvement from training on the full category set $\mathcal{C}$ versus only on $c_t$ satisfies $\Delta \text{Gen} \geq \sqrt{\frac{d_{\text{model}}}{|\mathcal{D}{c_t}|}} - \sqrt{\frac{d{\text{model}}}{|\mathcal{D}{\text{full}}|}} + \lambda \cdot R{\text{transfer}}$ where $c_t$ denotes the target category, $\mathcal{C}$ is the complete set of product categories, $|\mathcal{D}{c_t}|$ is the data size of $c_t$, $|\mathcal{D}{\text{full}}| = \sum_{c \in \mathcal{C}} |\mathcal{D}c| \gg |\mathcal{D}{c_t}|$ is the total cross-category data size, $d_{\text{model}}$ is the model complexity, $\lambda$ is a coefficient, and $R_{\text{transfer}}$ represents the regularization effect from cross-category knowledge transfer.

\textit{Proof Sketch.} The bound follows from comparing the standard generalization bounds for single-category training (which suffers from data scarcity) and cross-category training (which benefits from aggregated data). The additional $R_{\text{transfer}}$ term captures how semantically similar categories provide complementary information that regularizes the model. The full proof uses information-theoretic arguments to formalize this intuition.

\section{Experiments}

\subsection{Dataset Construction}

\subsubsection{Dataset Construction}

\begin{figure}[t]
\centering
\includegraphics[width=0.96\columnwidth]{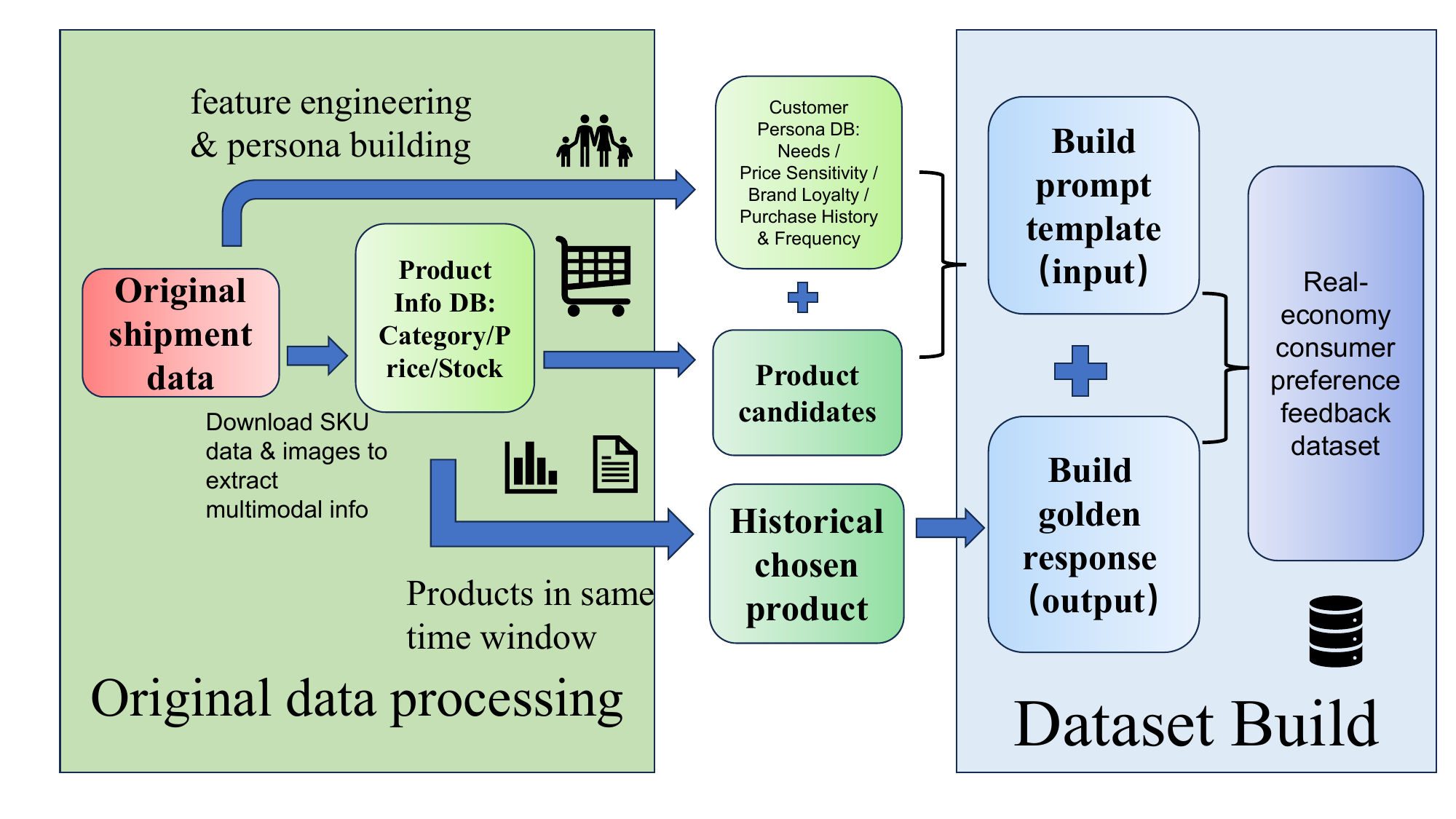} 
\caption{Dataset building pipeline from real trade data.}
\label{fig:ved_sr}
\end{figure}

As shown in \ref{fig:ved_sr} MALLES economic sandbox dataset embeds real-world sales data, including product choices and procurement quantities, into structured prompts. Each prompt integrates customer demand, candidate product details (multimodal, including pricing), and historical purchasing behavior to capture individual decision tendencies. Although data for individual customers or specific categories can be sparse, aggregation over $119252$ customers and $3361$ distinct categories yields a comprehensive representation of real-market decision patterns.

Historical sales records are organized into procurement selection prompts for single-agent LLMs. Each prompt contains \textbf{[competitive product information and pricing, historical purchase records and prices, market purchasing trends, review ratings, promotional offers]}, guiding the model to generate \textbf{[purchase selection, purchase quantity]}. Model interaction is implemented via \texttt{llm\_chat(prompt)}. We construct the single-agent alignment dataset, where \texttt{input} and \texttt{output} fields correspond to prompts and historical purchase decisions, serving as alignment data for the commercial simulation model.

\subsubsection{Experimental Motivation and Design}
Retail and wholesale decision-making is inherently high-dimensional, multimodal, and data-sparse. Traditional rule-based or shallow learning approaches often exhibit unstable predictions and poor interpretability when facing combinatorial action spaces and cross-category transitions. Our framework addresses these by post-training on large-scale transaction data with injected numerical sensitivity mechanisms, enabling LLMs to develop monetary awareness and numerical reasoning for accurate simulation. In parallel, multi-agent gaming and symbolic regression loops in wholesale scenarios allow the model to discover actionable procurement rules through linguistic interaction, balancing strategy quality with interpretability.

We validate MALLES in retail scenarios under multiple configurations including model choice (base vs. post-trained), decision framework (multi-agent multi-round discussion vs. single-agent default), and input sampling (mean-field vs. standard). Mean-field sampling is not applied in multi-agent discussions. We sample the bottom 50\% of customers by historical purchase volume to represent ordinary consumers. In multi-agent settings, several agents take turns over multiple rounds before producing a summarized decision, whereas single agents generate outputs directly. Post-trained and base models differ only in the model checkpoint, and mean-field sampling replaces scattered samples with averaged field observations in single-agent scenarios.

For all LLMs inferences in this section, including data construction, sandbox simulation and evaluation, we employ four representative models: GPT5.2\cite{gpt5}, Gemini3\cite{gemini}, DeepSeek-V2\cite{deepseek-v2} and Llama-4(70B)\cite{llama}. For each inference one model is randomly selected to generate result the response. For evaluation task with scalar outputs we use the average of parsed scores as the final metric.

\subsection{Simulation and Evaluation}

\begin{figure}[t]
\centering
\includegraphics[width=0.96\columnwidth]{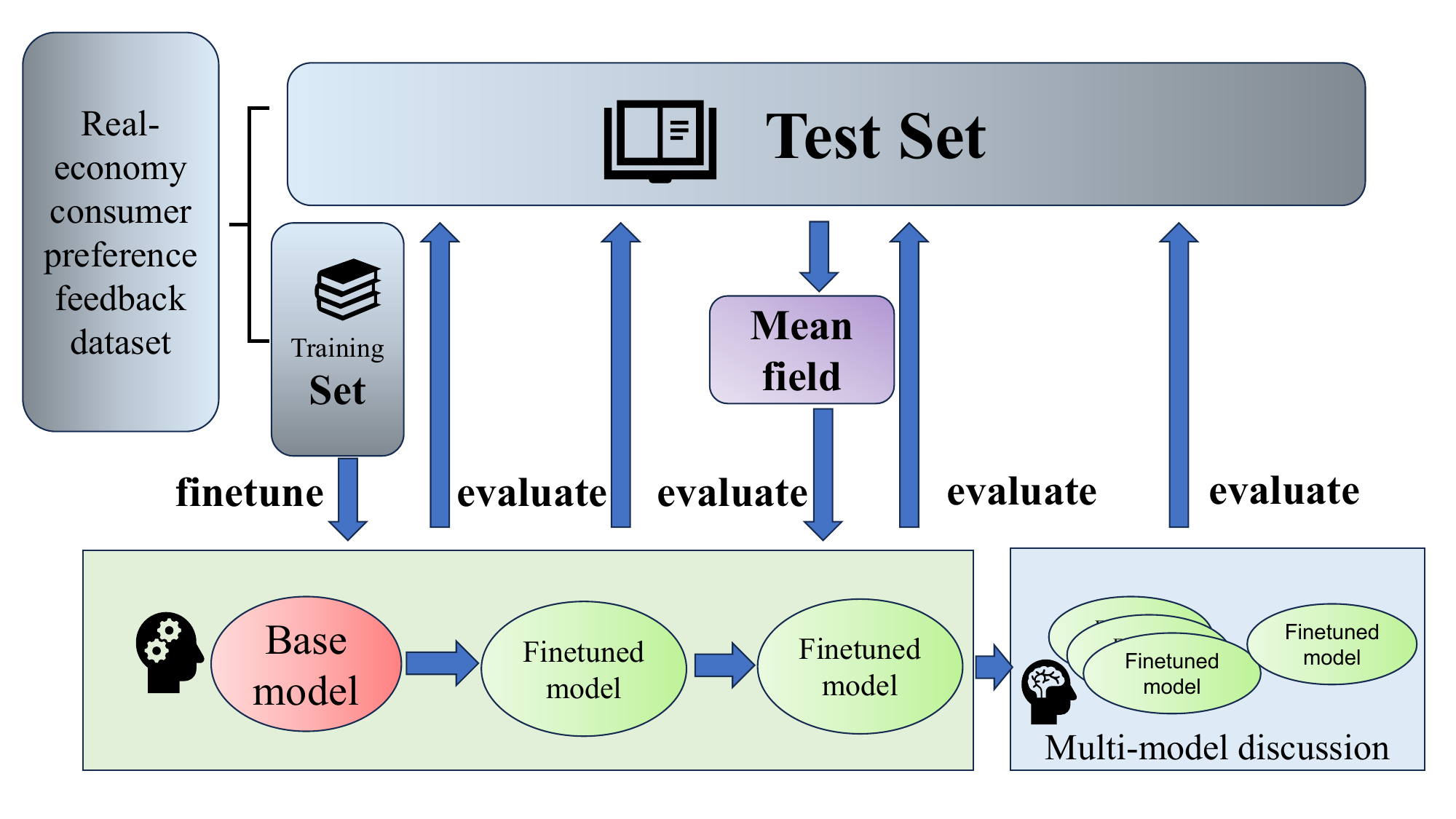} 
\caption{Major simulation pipeline of introduced sandbox.}
\label{fig:pipeline}
\end{figure}

\textbf{Setup.} As \ref{fig:pipeline} display the retail experiment evaluates whether post-trained LLMs can learn stable price preferences and purchase patterns from real transaction data while retaining reasonable out-of-distribution generalization. Following consumer decision theory, model inputs include customer profiles, candidate product features (text and image embeddings), and discount information. The model reasons over these inputs to produce purchase intent and quantity predictions. We compare against four baseline LLM-based economic simulators to assess whether LLMs can achieve intrinsic numerical alignment in economic behavior.

We evaluate three dimensions: (1) responsiveness to price and discount changes, measured by SKU selection accuracy and purchase quantity consistency against historical records; (2) prediction quality on out-of-distribution categories, including unseen brands or products; and (3) stability, quantified by the variance of predicted purchase quantities across multiple samples (lower variance indicates higher stability).

The experiment uses real industrial sales data, sampling 1,000 customer-product-quantity instances. Given the context, agents act as consumers to select preferred products and predict order quantities. Predictions are compared with historical outcomes, and candidate orders are randomly shuffled to mitigate positional bias, particularly for items appearing first in the list.


\begin{figure*}[t]
\centering
\includegraphics[width=1.7\columnwidth]{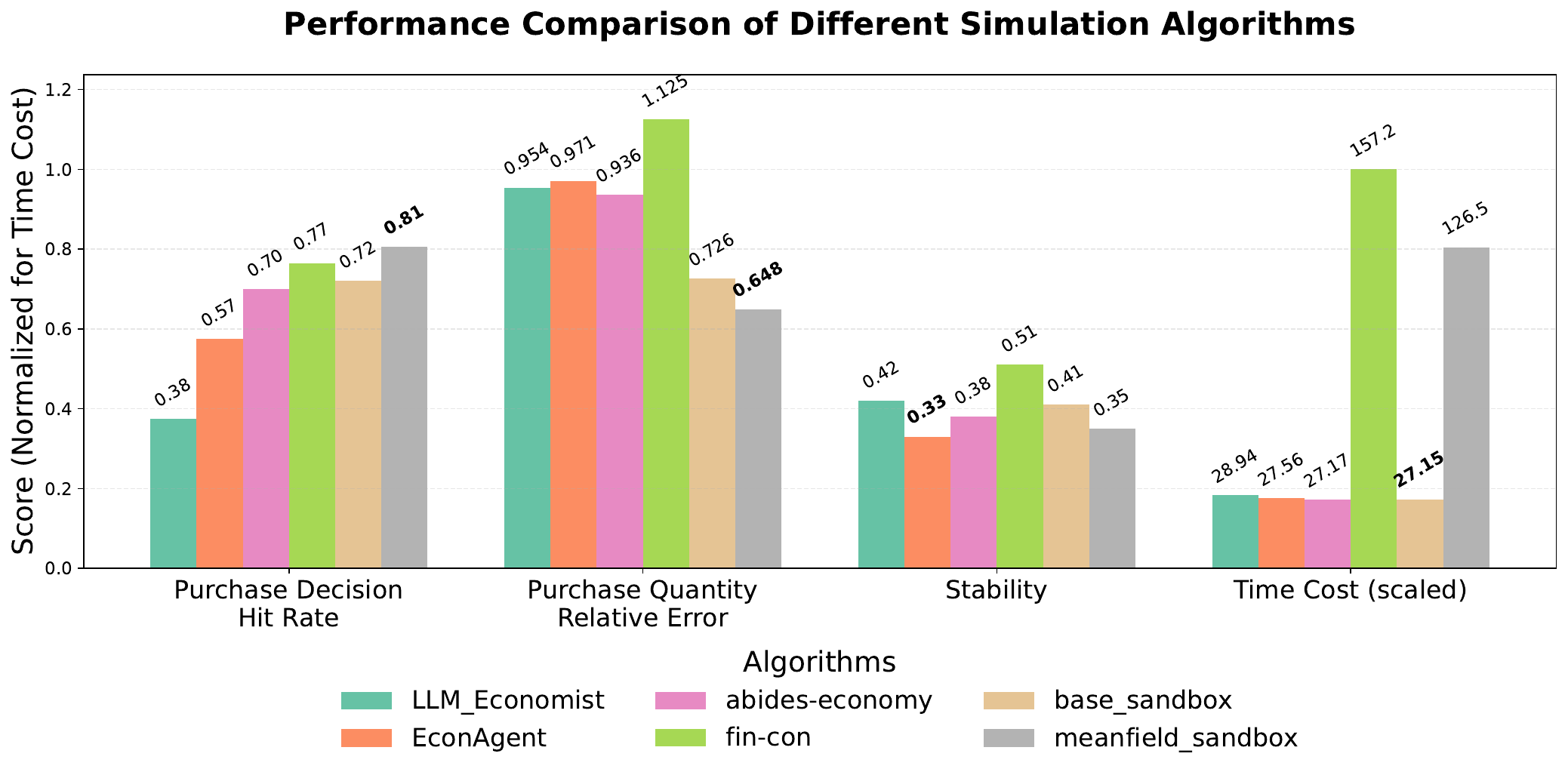} 
\caption{Performance comparison of different simulation algorithms. The best one  in each metric is \textbf{bolded}. Time Cost values are normalized to the [0,1] range with original values on the bars. The base\_sandbox and meanfield\_sandbox collectively correspond to the MALLES approach with and without mean-field enhancement, respectively.}
\label{fig1}
\end{figure*}

\textbf{Results Analysis.} Results \ref{fig1} demonstrate that LLM-based economic sandbox MALLES achieves consistent advantages in consumer behavior simulation. In purchase decision hit rate, both configurations outperform existing baselines, with the basic version reaching 0.700 and the enhanced version 0.775. This indicates that LLMs effectively learn consumer preferences via cross-category training, overcoming the limitations of traditional methods in high-dimensional settings. While fin‑con attains a slightly higher hit rate (0.800), its substantially large quantity error (1.325) and lower stability reveal systematic biases in quantitative prediction. Other baselines (e.g., abides‑economy) maintain controlled errors but exhibit limited hit rates, highlighting the constraints of rule‑based or shallow‑learning approaches.

MALLES maintains moderate quantity prediction errors while achieving high hit rates, reflecting the benefits of improved numerical sensitivity induced by monetary awareness instilled during post‑training. Overall, the system provides the best balance between decision accuracy and quantity error, with stability metrics surpassing most baselines. This verifies that injecting numerical sensitivity through post‑training enables LLMs to learn price elasticity and preference transition patterns more effectively.

Compared to baselines, MALLES excels in superior stability benefiting from attention weight control and consistency regularization that reduce output variance in high‑dimensional decision spaces. From a computational perspective, the base version incurs costs comparable to most baselines, while the enhanced version trades increased overhead for substantial performance gains. These results demonstrate that combining LLMs’ semantic reasoning with economic structure enables high‑fidelity simulations that preserve individual heterogeneity while aligning with macro‑market patterns, offering a reliable foundation for real‑economy decision‑making.

\subsection{Ablation Studies}

\begin{figure*}[htbp]
\centering

\includegraphics[width=0.39\textwidth]{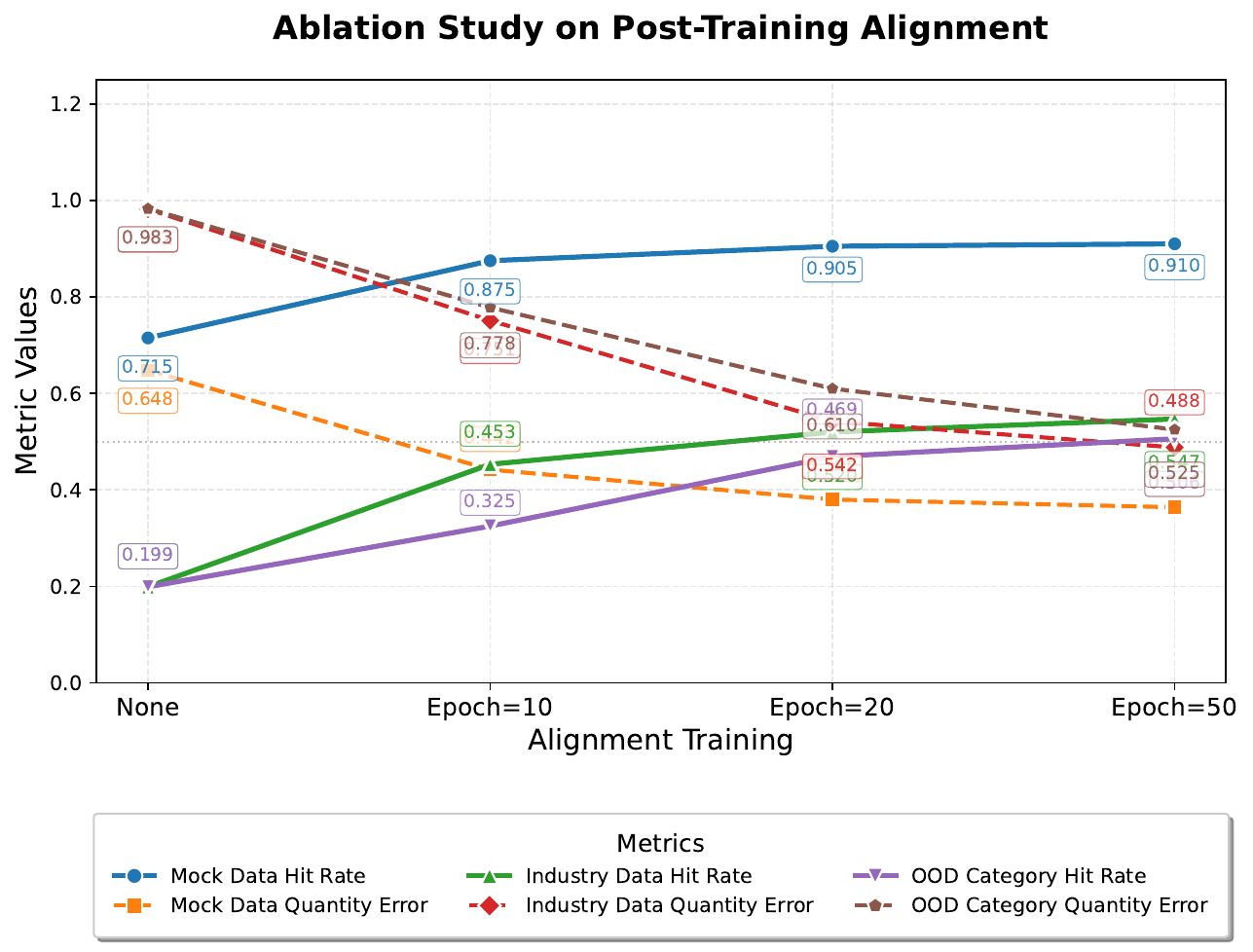}
\hfill
\includegraphics[width=0.30\textwidth]{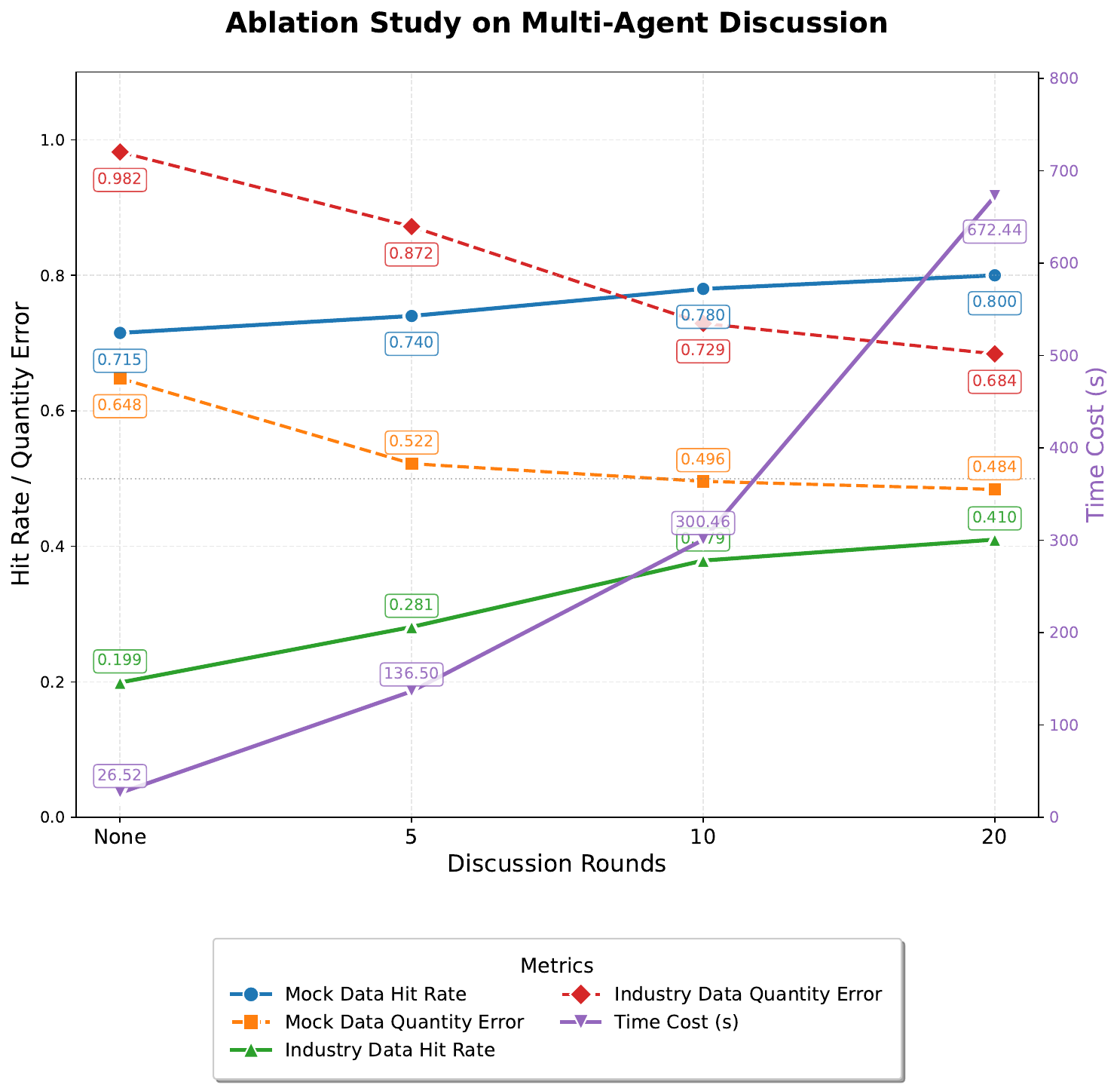}
\hfill
\includegraphics[width=0.30\textwidth]{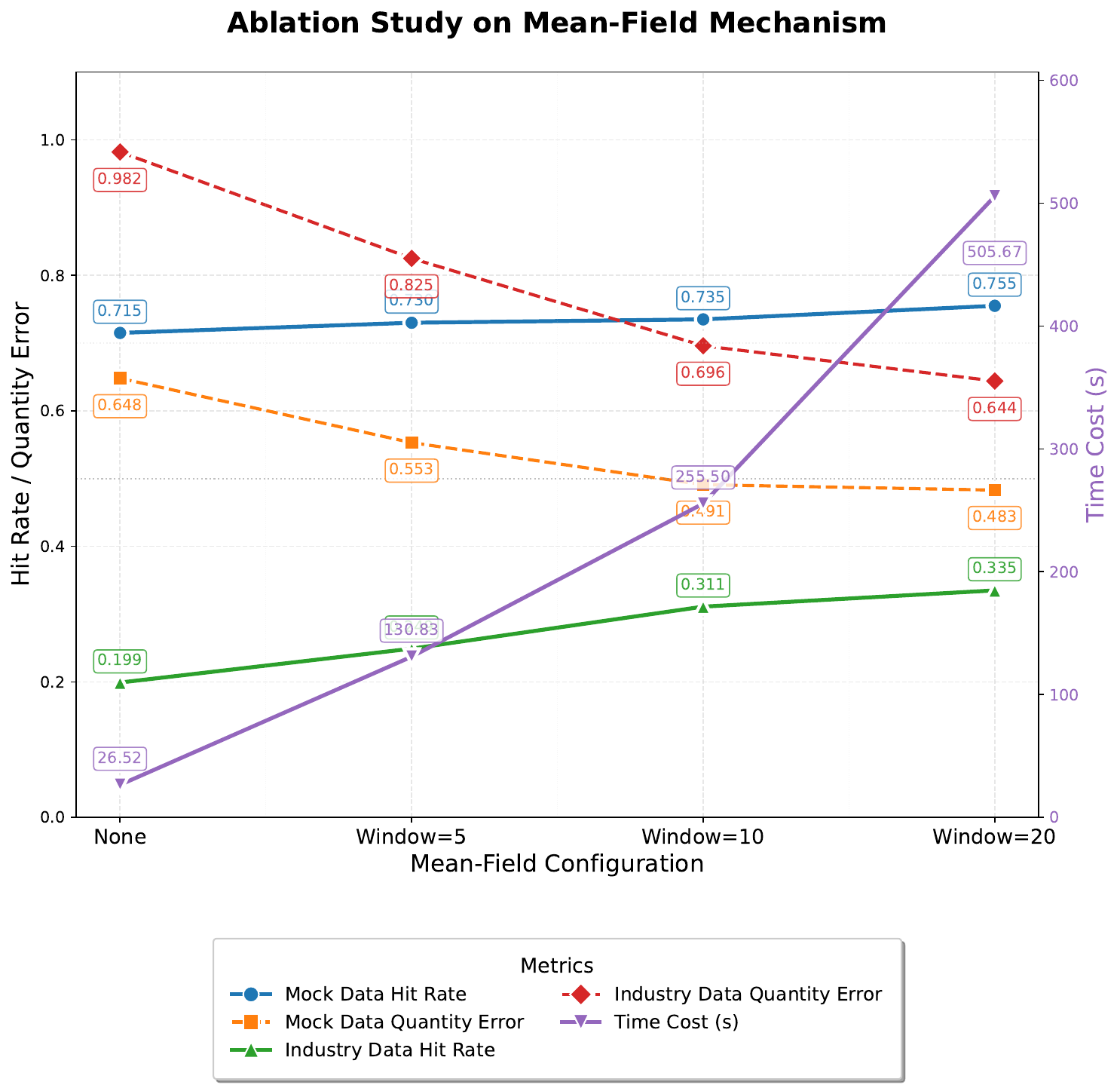}

\vspace{2pt}

\textbf{(a) Ablation on post-training alignment.}\label{fig:sub1}
\hfill
\textbf{(b) Ablation on multi-agent discussion}\label{fig:sub2}
\hfill
\textbf{(c) Ablation on mean-field mechanism}\label{fig:sub3}

\caption{Ablation study on multiple objectives}
\label{fig:ablation_study}
\end{figure*}

We conduct systematic ablations on three key mechanisms: (1) post‑training alignment, (2) multi‑agent discussion, and (3) mean‑field observation. Each experiment modifies only the target module while keeping other configurations consistent to isolate its contribution. Results are shown in \ref{fig:ablation_study}.

\textbf{Post-training vs. No Post-training.} Training explicitly teaches price elasticity, discount sensitivity, and cross‑category preference transitions, aligning both product selection and quantity estimation with observed consumer behavior. Removing this stage leaves the model to rely solely on pre‑trained linguistic priors, increasing the risk of discontinuous price responses and distorted numerical judgment.

We construct four control groups based on training status and training epochs. Metrics include selection hit accuracy, purchase quantity error(as defined in the main experiment), and hit rate on OOD categories unseen during training. All experiments are conducted under default conditions without mean‑field or multi‑agent enhancements, using two test sets: held-out real industry transaction data and a mock dataset abstracted from historical interactions via stepwise style summarization. We sample 1,000 test instances per group. With SFT (1,024 samples per epoch), models are trained for 10, 20, and 50 epochs, repeating N=10 times and  final metrics are averaged. For OOD evaluation, models trained on ‘Home Cleaning’, ‘Paper Products \& Wipes’, and ‘Laundry Detergent \& Care’ are tested on ‘Daily Necessities’, ‘Laundry Cleaning’, and ‘Cleaning Paper Products’.


The ablation results confirm the critical role of post-training alignment. The base LLM shows limited selection commonsense (~0.2 hit rate on industry data). As training epochs increase, hit rates improve and quantity errors decrease, indicating the LLM learns progressively learns economic regularities like price elasticity and discount sensitivity. In OOD testing, post-trained models show strong generalization with gains in quantity prediction accuracy, verifying that cross-category training yeilds transferable preference representations that mitigate data sparsity.

Notably even with limited epochs, models achieve substantial gains, indicating high data efficiency. Although performance on mock data exceeds that on real industry data—reflecting additional noise and latent variables—aligned models consistently outperform random baselines and provide valuable decision guidance. Post-training alignment converts LLMs' semantic understanding into robust economic numerical economic rasoning, forming a reliable bais for digital economic sandboxes.

\textbf{Single-Agent vs. Multi-Agent.} In real-world strategy optimization, decisions often  emerge from interactions among multiple roles (e.g., manufacturers, distributors and customer service). Our multi-agent design aims to expand strategy exploration through role specialization and stochastic interaction, reducing risk of single‑agent local optima. We investigate whether single-agent systems can achieve comparable benefits under matched computational budgets.

We compare two systems: (a) a multi-agent framework with 3–4 collaborative roles exchanging observations and objectives via linguistic interaction; and (b) a  single‑agent framework where all decision logic is handled by a single model instance with equivalent compute computational resources and environment, but without role interactions. Both operate in wholesale simulation until strategy convergence, using identical profit, inventory turnover, and evaluation metrics. This ablation isolates the effect of role interaction on strategy diversity, convergence behavior, and “division of labor and gaming” in avoiding local optima and promoting rule discovery.



Results show role specialization and interaction improve decision quality, though marginal gains diminish as the number of discussion rounds increasese, while computational cost grows linearly.  With few discussion rounds, multi‑agent systems offer limited advantage over single‑agent one, underscoring value of multi‑perspective discussion primarily in avoiding local optima. With more rounds, the system better integrates domain-specific insights for more comprehensive strategies.

The multi‑agent framework compresses context via differentiated input partitioning, distributing cognitive load across agents. However, improvements in purchase quantity prediction are smaller, suggesting that numerical precision may require additional targeted training, potentially due to limited quantity-specific background knowledge. Compared to single-agent systems, multi‑agent ones better capture critical decision factors through structured dialogue at reasonable cost, making them particularly suitable for complex wholesale scenarios, while single-agent systems remain preferable for efficiency-sensitive retail applications.

\textbf{Effect of Mean-Field on Decision Simulation Stability and Market Response Distribution Error. }For single‑agent procurement simulation, we examine the effect of mean‑field processing by incorporating moving‑window averages of historical market responses as environmental context within prompts built referred in methodology section, while keeping other configurations unchanged. The window unit is statistical months.

\begin{table*}[t]
\centering
\caption{Ablation study on cross-category training}
\begin{tabular}{lccccc}
\hline
Training Data & Background & Single-Category & Mixed & Similar-Category & Full-Category \\
\hline
\textbf{Paper Wipes} & & & & & \\
Hit Rate & 0.25 & 0.56 & 0.32 & 0.50 & 0.48 \\
Quantity Error & 0.962 & 0.240 & 0.473 & 0.399 & 0.425 \\
Stability & 0.70 & 0.45 & 0.65 & 0.62 & 0.62 \\
\hline
\textbf{Home Cleaning} & & & & & \\
Hit Rate & 0.24 & 0.60 & 0.49 & 0.57 & 0.55 \\
Quantity Error & 0.988 & 0.482 & 0.581 & 0.527 & 0.539 \\
Stability & 0.66 & 0.61 & 0.62 & 0.65 & 0.69 \\
\hline
\textbf{Laundry Detergent} & & & & & \\
Hit Rate & 0.15 & 0.61 & 0.42 & 0.55 & 0.53 \\
Quantity Error & 0.993 & 0.492 & 0.467 & 0.597 & 0.561 \\
Stability & 0.69 & 0.66 & 0.61 & 0.59 & 0.62 \\
\hline
\end{tabular}
\label{tab:ablation-cross}
\end{table*}



Incorporating historical response statistics improves decision accuracy and stability for single-agent models. As observation window length increases, both purchase decision hit rate and quantity prediction accuracy improve while output variance controlled. This indicates that combining micro-level individual decision cues with macro-level population statistics mitigates biases from incomplete information. Medium-length windows capture most improvements, with longer windows yielding diminishing returns.

Compared with using scattered historical samples directly, mean-field processing provides more representative market background, which is particularly beneficial for price-sensitive decisions. By introducing population-level statistical corrections, this mechanism compensates for information gaps in individual decisions, aligning with the input enhancement principles described in methodology. In practice, properly configured mean-field mechanisms significantly improve simulation robustness and reliability without excessive computational overhead, making them well suited for real-world scenarios requiring rapid market response.

\textbf{Cross-Validation of Full-Category Data for Enhancing LLM Prediction Accuracy. } Training with full-category sales data further leverages LLMs' semantic generalization abilities—such as capturing consumer preference patterns and naming conventions—to enhance alignment. This approach is more scalable than traditional methods, which require manually designed cross-domain mappings under homogeneous distribution assumptions. Crucially it enables effective utilization of large-scale cross-category and cross-domain data to improve preference alignment, particularly for data-scarce categories or entirely new products, thereby alleviating out-of-distribution challenges common in conventional prediction systems.

We validate across multiple product categories, training models for 10 epochs under different data compositions: alpaca data (background), category-specific historical data, mixed historical and alpaca data, similar-category data (same first-level category), and full-category data. Evaluation metrics remain SKU hit rate, purchase quantity relative error, decision stability, and time cost.

The cross-category training results \ref{tab:ablation-cross} highlight the unique value of full-category data for consumer preference learning. Across test categories, models trained on full-category or similar-category data significantly outperform those trained only on background or mixed background–category data which reflects scenarios with limited labeled data. Although these models slightly underperform specialized single-category training due to reduced same-category data proportion, they provide a strong and practical alternative in data-sparse settings, such as new product launches, where full-category or near-category generalized models can provide reliable decision references.

Full-category aligned models maintain consistency across categories with balanced stability metrics, indicating consumer preference representations learned from heterogeneous data generalize effectively. Compared with traditional methods requiring complex feature engineering, LLMs naturally extract transferable consumption patterns from diverse data sources. Gains in selection hit rate are more pronounced than reductions in quantity error, reflecting LLMs' higher sensitivity to textual information than numerical signals. Results support hierarchical data strategies: leveraging category-specific fine-tuning in data-rich categories and relying on full-category training in data-limited scenarios.

\section{Conclusion}

This research introduces a novel economic simulation paradigm by integrating multimodal understanding of LLMs, multi-agent collaborative decision-making, and principles of economic systems. To address the core challenges of category-level data sparsity, poor out-of-distribution (OOD) generalization, and high-dimensional product feature complexity outlined in the introduction, we propose the Multi-Agent Large Language Model-based Economic Sandbox (MALLES). It leverages cross-category transaction data for economic alignment via post-training, enabling LLMs to learn and transfer underlying consumer preference patterns, thereby mitigating data scarcity.

Key methodological contributions include a multi-agent discussion mechanism for distributing cognitive load over long contexts, mean-field stabilization for modeling dynamic market interactions, and input augmentation with attention control for completing partial observations. They collectively enhance simulation fidelity, stability, and interpretability. Extensive experiments demonstrate that our approach significantly improves product selection accuracy, purchase quantity prediction, and simulation stability over existing economic LLM simulation baselines. Ablation studies further validate necessity of post-training alignment, multi-agent discussion, and mean-field mechanisms within our proposed framework.

\section{Limitations and Ethical Considerations}

Several promising directions remain. Deeper investigation into cross-modal alignment is warranted, particularly for more effectively fusing visual, textual, and numerical product information into unified economic decision representations, thus addressing current limitations in modal integration. Achieving a tighter coupling between micro-level agent decisions and emergent macro-level market dynamics remains an open challenge. While our mean-field method provides a connection, more sophisticated multi-scale modeling is needed to capture complex network effects and supply chain interdependencies in hierarchical economic systems. For ethical concern, all personal information in training data should be rigorously anonymized and aggregated to prevent re‑identification, and capacity for accurate behavioral prediction raises risks over the potential design of manipulative patterns. Ethical guidelines must be enforced to prevent overly personalized tactics that could undermine consumer autonomy.

\bibliography{aaai22}

\end{document}